# Evidence for systematic semantic structure in individual letters

Gexin Zhao[1,*]
[1]*Columbia University, New York, NY, USA.*

**Corresponding author.*
*Email: gz2393@cumc.columbia.edu*

**Abstract**

Associations between speech sounds and meaning are well documented but have not been systematically mapped over a whole alphabet. Here we map them across the 26 English letters and find that each carries a structured, multidimensional semantic profile that is recoverable from text, perceived across languages, and predicted by articulatory features. Three large language models independently detected consistent semantic structure across nine perceptual dimensions in 220 pairwise letter contrasts, and the profiles they recovered were then tested in preregistered experiments with 1,388 human participants. Native English readers chose the predicted word above chance (85.3%, selected items; 65.7%, all contrasts), and the preference followed the letter rather than the words that carried it. Listeners of five typologically diverse languages showed the same preference (76.7%, selected pairs; 68.4%, randomly drawn contrasts), regardless of their English proficiency. Articulatory features assigned to each letter predicted both the model profiles and the human judgments. Letter–meaning association is thus a systematic, multidimensional property of the alphabet rather than a set of isolated effects.

## Introduction

Whether the sounds of a word bear any relation to its meaning has been debated since Saussure's principle of the arbitrary sign[1], and classical phonology treats phonemes as units that distinguish words rather than carry meaning of their own[2]. However, that description has been contested for nearly a century: Sapir[3] showed in 1929 that listeners take a nonsense word with "***a***" to name a larger object than the same word with "***i***", Bolinger argued in 1949 that the sign is not arbitrary[4], and later work argued that many sound–meaning pairings are not accidental, and occur across unrelated languages[5–8]. Such sound–meaning correspondences, from bouba/kiki to vowel–size mappings, are now known to be widespread in natural-language vocabularies[9–12].

Several lines of evidence suggest that such correspondences are more basic and more widespread than an occasional bouba/kiki effect. They recur across thousands of languages after historical relatedness and contact are controlled for[13]. They persist across writing systems[14], emerge in non-human species without linguistic experience[15], and are retained by deaf and hard-of-hearing individuals[16], which suggests that the experience of producing a sound, not only hearing it, can ground its meaning. They are also one case of a broader tendency to match features across the senses[17,18], which in part reflects regularities of the physical world.

If these correspondences are that basic, they should not be confined to the selected subsets studied so far[3,19–21], but should run through the whole inventory from which words are built. In this study, we ask this of the 26 English letters: whether each letter carries a distinct perceptual meaning. If so, whether that meaning is perceived by listeners of other languages who never see the spelling, and if it is, where it comes from.

Four studies address these questions: a mapping of all 220 pairwise letter contrasts on nine dimensions with language models as the rating instrument (Study 1), preregistered forced-choice tests with English readers over the complete set of contrasts (Studies 2a and 2b) and with listeners of five languages hearing the nonwords spoken (Studies 3a and 3b), and an articulatory-feature analysis of the resulting profiles (Study 4).

## Results

### Study 1: Language-model ratings indicate multidimensional structure in letter contrasts

To map the landscape of letter–meaning associations, we constructed 4,420 minimal nonword pairs spanning all 220 pairwise letter contrasts (~20 pairs per contrast, Fig. 1A). In each pair, two nonwords differed by exactly one target letter while all frame letters were held constant (e.g., brev/brov, contrasting e vs. o; see Methods). Three large language models (GPT-4o, Claude Sonnet 4, and Gemini 2.5 Flash Lite) independently rated every nonword on nine semantic dimensions: size, shape (round/spiky), brightness, texture (smooth/rough), speed, temperature, pleasantness, weight, and elevation (Table S1). The nine dimensions were chosen to span the perceptual domains most consistently implicated in the prior sound-symbolism literature[11,22], including size, shape, brightness, texture, and pleasantness, and were fixed before any data collection.

Every letter carried a distinct, multidimensional semantic profile (Fig. 1B). Aggregating pair-level Cohen's d values to the letter level revealed systematic patterns: ***i*** was associated with smallness, brightness, speed, and lightness, whereas ***o*** was associated with largeness, darkness, and heaviness. The nine dimensions showed structured covariation (Fig. 1C; full matrix in Fig. S1D): size and weight were strongly correlated ($r = 0.95$), as were brightness and elevation ($r = 0.85$). Pleasantness was inversely related to texture/roughness ($r = –0.95$) and moderately to weight ($r = –0.65$).

Pooling across three models, the mean absolute Cohen's d was 0.59, with 79% of effects at least small in size ($|d| > 0.2$) and 51% medium or large ($|d| > 0.5$). Of 220 letter pairs, 218 (99%) produced significant differentiation on at least one dimension, and 87% (192) differentiated on three or more dimensions simultaneously (Dataset S1). The differentiation is far above chance (permutation test, $p < 0.001$; Methods; Dataset S1) and stable across models (Fig. S1A). Split-half reliability, assessed over 100 random splits of nonword carriers, yielded Spearman-Brown corrected r values of 0.79 (temperature) to 0.92 (weight) across dimensions and models (Fig. S1B), confirming that the observed effects reflect the target letter contrast itself rather than the particular carrier words. The effect also scales with dose: pairs in which the target letter occurs twice show 1.2–1.5 times the effect of single-occurrence pairs on every dimension (Fig. S1C), as expected if each occurrence adds to the nonword's profile.

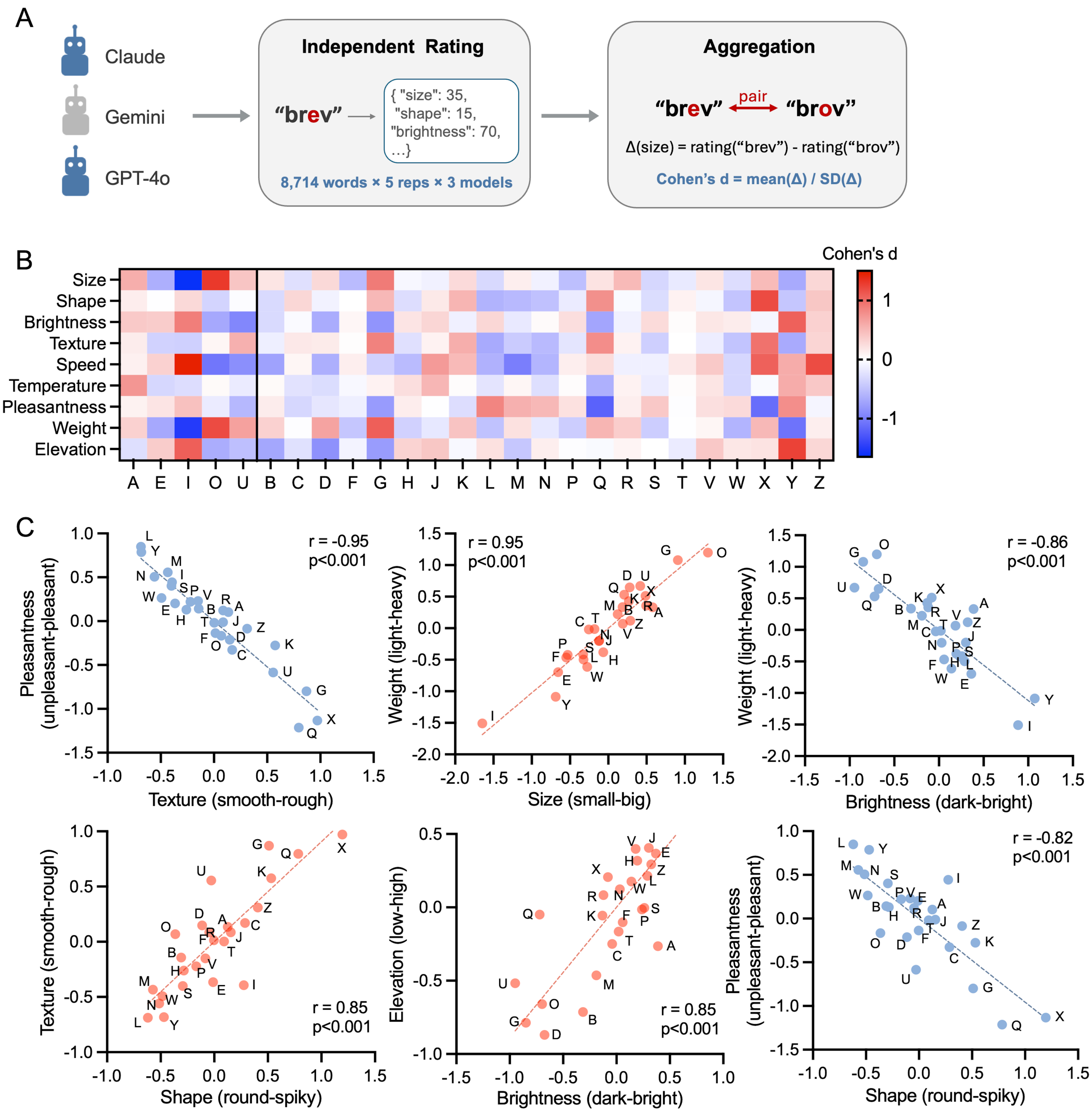

**Figure 1. Letter-level semantic profiles and the covariation among dimensions.**
(A) Rating procedure. Claude Sonnet 4, Gemini 2.5 Flash Lite, and GPT-4o each rated 8,714 nonwords on nine dimensions, five times per word. For each minimal pair (e.g., brev vs. brov), the two words' ratings were subtracted; Cohen's d for a letter contrast is the mean of these differences over all pairs with that contrast, divided by their SD.
(B) Mean signed Cohen's d for each letter on the nine dimensions; vowels (A, E, I, O, U) on the left, consonants on the right. Red, upward shift on the dimension; blue, downward shift. |d| > 0.5 denotes a medium effect and |d| > 0.8 a large effect.
(C) Selected inter-dimension correlations across the 26 letters. Each point is one letter; dashed lines, OLS fits with Pearson r and p. Red panels, positive correlation; blue, negative.

**Study 2: Preregistered forced-choice tests with English readers over all 220 contrasts.** To test whether human participants detect the same letter–meaning associations identified by LLMs, we conducted preregistered forced-choice experiments with native English speakers (Methods). On each trial, participants saw two nonwords and chose which better matched a given semantic pole (e.g., "Which sounds bigger?"; Fig. 2A). Study 2a tested 54 nonword pairs, six per dimension, selected from the Study 1 pool for high and consistent model effects (Table S2). Participants (N = 93) chose the LLM-predicted nonword on 85.3% of trials, well above chance on eight of the nine dimensions (82.6–94.6%; Fig. S2A). Temperature was the weakest dimension for participants (55.2%), as it was for the models. Apart from three temperature items, the majority choice agreed with the model on every item (Fig. S2B).

To ask whether this generalizes beyond selected items, Study 2b tested all 220 contrasts with carrier words drawn at random. Each contrast was tested on the dimension with its largest LLM effect with two nonword pairs (N = 688; Fig. 2B–G; Table S3). Agreement fell to 65.7% (95% CI [65.0, 66.4]) but remained above chance on eight of the nine dimensions (Fig. 2D–E). The majority choice agreed with the model on 378 of the 440 items and on 191 of the 220 contrasts, and agreement scaled with the size of the LLM effect (Fig. 2B). Direction and size of the model effect together predicted the human choice across the 440 items (Fig. 2C).

The two carrier pairs of each contrast also show whether the effect follows the letter or the words that carry it. The proportion of participants choosing the word with the alphabetically earlier letter correlated at $r = 0.83$ between the two pairs ($p = 5.1 \times 10^{-58}$; Fig. 2F), whereas the difference between the two carriers' model ratings predicted the difference in human choice only weakly ($r = 0.27$; Fig. 2G). Human judgments therefore depend mainly on the letter contrast, with the particular carrier words shifting the preference by a median of 7 percentage points.

Response times fell as item agreement rose (Fig. S2D–E), and participants were faster when they chose the predicted word (Fig. S2F). At the item level, consistent responses became faster as agreement rose while inconsistent responses did not (Fig. S2G), a pattern that fits a quick, intuitive judgment rather than deliberation. Accuracy was stable across trials while response times fell (Fig. S2H–I), so the judgments were available from the outset rather than learned during the task.

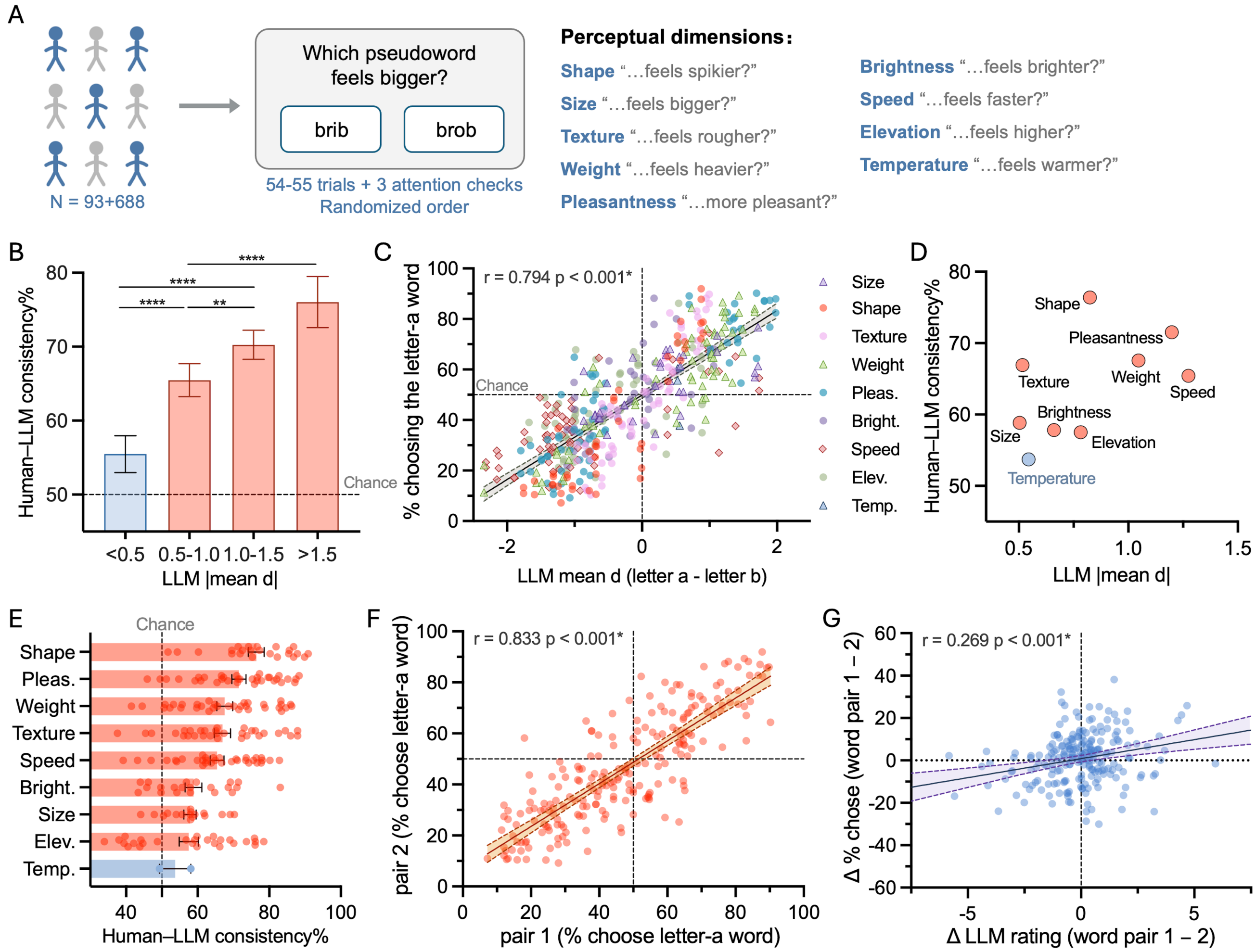

**Figure 2. Preregistered forced-choice tests with English readers over all 220 contrasts**
(A) Participants (Study 2a, N = 93; Study 2b, N = 688) saw a question and two nonwords and chose one: 54 (2a) or 55 (2b) trials plus three attention checks, in random order. Right, the question asked for each dimension.
(B–G) Study 2b. (B) Percentage of participants choosing the LLM-predicted word, by the size of the model effect for the contrast (|mean d|; n = 100, 158, 140, 42 items per band). (C) Model effect for the contrast on the dimension tested (mean d, letter a − letter b) against the percentage choosing the letter-a word, all 440 items; color, dimension. (D) The same at the dimension level (nine points), descriptive only. (E) Consistency by dimension; points, contrasts (the two carrier pairs averaged). Blue, not above chance (temperature, two contrasts, four items). (F) Carrier generalization: for each of the 220 contrasts, the percentage choosing the letter-a word in the first and in the second independently drawn carrier pair (r = 0.83). (G) Within-contrast differences: the difference between the two carriers' model ratings against the difference in the human percentage (r = 0.27).
Bars, mean ± 95% CI (B, E); brackets, pairwise comparisons (** $p < 0.01$, **** $p < 0.0001$); solid lines, OLS fits with 95% CI, r = Pearson correlation (C, F, G).

**Study 3: Preregistered auditory experiments with listeners of five languages**

Study 2 used written nonwords presented to English speakers, leaving open whether the observed effects depend on English orthographic or visual features. To dissociate acoustic information from orthography, we conducted a follow-up study in which the nonwords were presented as synthesized speech to native speakers of English, Spanish, Japanese, Mandarin, and Arabic, who never saw the spelling (Fig. 3A; Methods).

Study 3a used 30 pairs selected from Study 1 for high model consensus and pronounceability (N = 155; 29–34 per language). Rather than the nine dimensions of Study 1, each pair was asked the question on which its two letters differed most, often a concrete attribute such as "sinking", "piercing", or “turning/revolving” (22 questions across 30 pairs; Table S4). All five groups chose the predicted nonword above chance, with median accuracies of 76.7% (English), 75.0% (Spanish), 83.3% (Japanese), 76.7% (Mandarin), and 73.3% (Arabic) (Fig. 3B), and 29 of the 30 pairs were above chance when pooled across languages (Fig. S3A).

Because those 30 pairs had been selected for high model consensus, Study 3b tested 40 further contrasts drawn at random from the 220, excluding the 30 of Study 3a. For 20 of them the nonword pair and its question were carried over from the print studies; for the other 20 the pair was drawn at random and the question written before data collection (Methods). Every pair was auditioned before data collection and rejected if the contrast was inaudible or if the audible difference extended beyond the contrast letters (Methods; N = 452; 18,080 trials). Mean accuracy was 68.4% (95% CI [67.6, 69.1]), all five groups were again above chance (Japanese 70.7%, English 70.3%, Mandarin 69.1%, Spanish 67.3%, Arabic 64.4%; Fig. 3B), and 38 of the 40 contrasts were above chance, 34 significantly so after correction; one (c–j) ran significantly against its registered direction (Fig. S3A and Table S5).

For the contrasts asked the same dimension question in print and in audio, the two agreed at $r = 0.92$ (Fig. 3C), and across all 70 auditory contrasts the models' choices also agreed with the listeners' ($r = 0.885$, $p = 2.5 \times 10^{-24}$; Fig. 3D). The five groups agreed with one another: the mean of the four non-native groups against English, $r = 0.93$ (Fig. 3E), and all ten pairwise correlations between groups $r = 0.79$–$0.89$ (Fig. 3F). Among the non-native groups accuracy did not vary with self-reported English proficiency (Fig. 3G). The 20 contrasts that had reached only 55.6% in print under a nine-dimension question reached 65.9% in audio under a question written for the pair (Fig. S3B), which suggests that a question written for the two letters captures their difference better than a fixed dimension does. As in Study 2, consistent responses became faster as agreement rose (Fig. S4A–B), and accuracy was stable across trials (Fig. S4C).

Rare letters do carry a flavor of their own, but it does not account for the effects (Fig. S4D–H). In Study 2b, when a contrast contained one of the six rarest letters (z, q, j, x, k, v), participants chose the rarer letter as the spikier (80.8%), rougher (68.8%), and faster (66.4%) one, and as the more pleasant only 29.5% of the time. For those contrasts with two common letters, no dimension showed a significant preference (Fig. S4D). In the auditory studies (Study 3), contrasts containing a rare letter were not more accurate than contrasts of two common letters (Fig. S4E). Adjusting each contrast for the difference in letter frequency shifted the human effect by at most 4.6 points and left every dimension above chance (temperature excepted, as it had only two contrasts; Fig. S4G), and how strongly a letter shifted the choices, in the human data as in the model ratings, was unrelated to its frequency (Fig. S4H; Dataset S2).

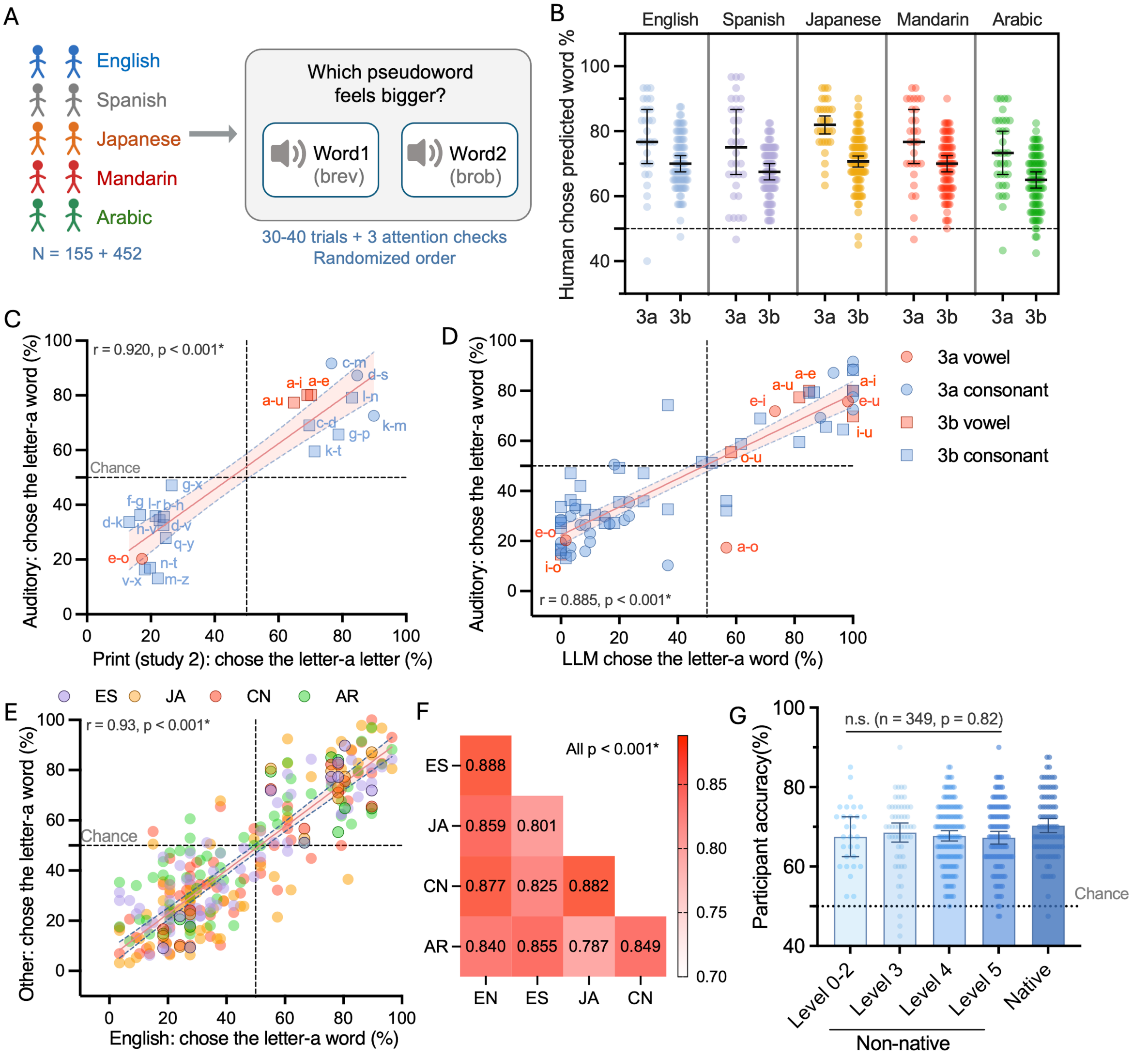

**Figure 3. Preregistered cross-linguistic auditory experiments (Studies 3a and 3b).**
(A) Native speakers of English, Spanish, Japanese, Mandarin, and Arabic (Study 3a, N = 155; Study 3b, N = 452) heard two synthesized nonwords and chose the one that better fit the question: 30 (3a) or 40 (3b) trials plus three attention checks, in random order, with the interface in each participant's native language.
(B) Percentage of trials on which participants chose the predicted word, by language and study; points, participants.
(C) The contrasts asked the same dimension question in print (Study 2b) and in audio: percentage choosing the letter-a word in each. (D) All 70 auditory contrasts against the models' forced-choice preference (percentage of model choices for the letter-a word). Circles, Study 3a; squares, Study 3b; red, vowel contrasts; blue, consonant contrasts.
(E–F) The four non-English groups against the English group, one point per contrast per group, with r computed between the English group and the mean of the four (E), and pairwise correlations between language groups over the 70 contrasts (F).
(G) Study 3b accuracy by self-reported English proficiency among non-native listeners (levels 0–2, 3, 4, 5) and for native speakers; points, participants.
Bars, median with 95% CI (B and G); solid lines, OLS fits with 95% CI, r = Pearson correlation (C–E); correlations in F are Benjamini–Hochberg corrected, all $p < 0.001$; G, one-way ANOVA across the non-native levels, $p = 0.82$.

**Study 4: Articulatory features predict the letter-level profiles**

Studies 1–3 established that letters carry multidimensional semantic signals detectable by both LLMs and humans across languages. Study 4 asked whether the semantic profiles can be predicted from each letter's articulatory features. Using the PanPhon feature system[23], we extracted articulatory features for the phoneme assigned to each letter (Methods) and tested whether the articulatory difference between two letters predicts their semantic difference, in the model ratings and in the human choices of Study 2b (Fig. 4A).

Feature–dimension associations were pervasive: for consonant contrasts, 91 of the 135 feature-dimension correlations survived correction at the pair level (Fig. S5), and the same correlations computed on the human choices closely matched those on the model ratings (Fig. S6A). Four examples are shown in Fig. 4B–C. Sonorant consonants read as rounder ($r = –0.64$ with shape) and smoother (–0.61 with texture), dorsal consonants (the feature back) as less pleasant (–0.61), and continuants as higher (0.45), and each of these relations reappeared in the human choices ($r = –0.64$, –0.60, –0.77, and 0.82). Among vowel contrasts, vowel height predicted size ($r = –0.97$), weight ($r = –0.95$), elevation ($r = 0.94$), and speed ($r = 0.88$), recovering the Sapir vowel–size gradient[3,24] (Fig. 4B, Fig. S6B).

Because each of the 21 consonants enters 20 of the 210 pairs, the pairs are not independent. We therefore repeated each test at the letter level ($n = 21$): the correlations keep their size, the four in Fig. 4B remain significant, though none survives correction over the 135 cells (Dataset S4). Letter frequency, which is correlated with the features (most strongly with sonorant, $r = 0.43$), does not explain these relations (Fig. S6C).

We then asked how well the features together predict the profile of a contrast the regression was not fitted on. Under leave-one-pair-out cross-validation the consonant regression reached $R^2$ = 0.56–0.84 across the nine dimensions (Fig. S5B). In the full fit, removing each feature class in turn showed manner of articulation contributing most to weight, size, and brightness, place to size, weight, and texture, and voicing to size and weight (Fig. S5C). Holding out a whole letter rather than a pair is a harder test, and accuracy fell (r from 0.87 to 0.49; $R^2$ from 0.75 to –0.16): the predictions still pointed the right way for most letters but missed their size, and the loss was concentrated on letters with no close counterpart among the 21, such as h, r, and l (Fig. S5D–E). Articulatory features thus predict the profiles of the letters they were fitted on, but the profile of an unseen letter cannot be fully recovered by adding up the effects of its features[21].

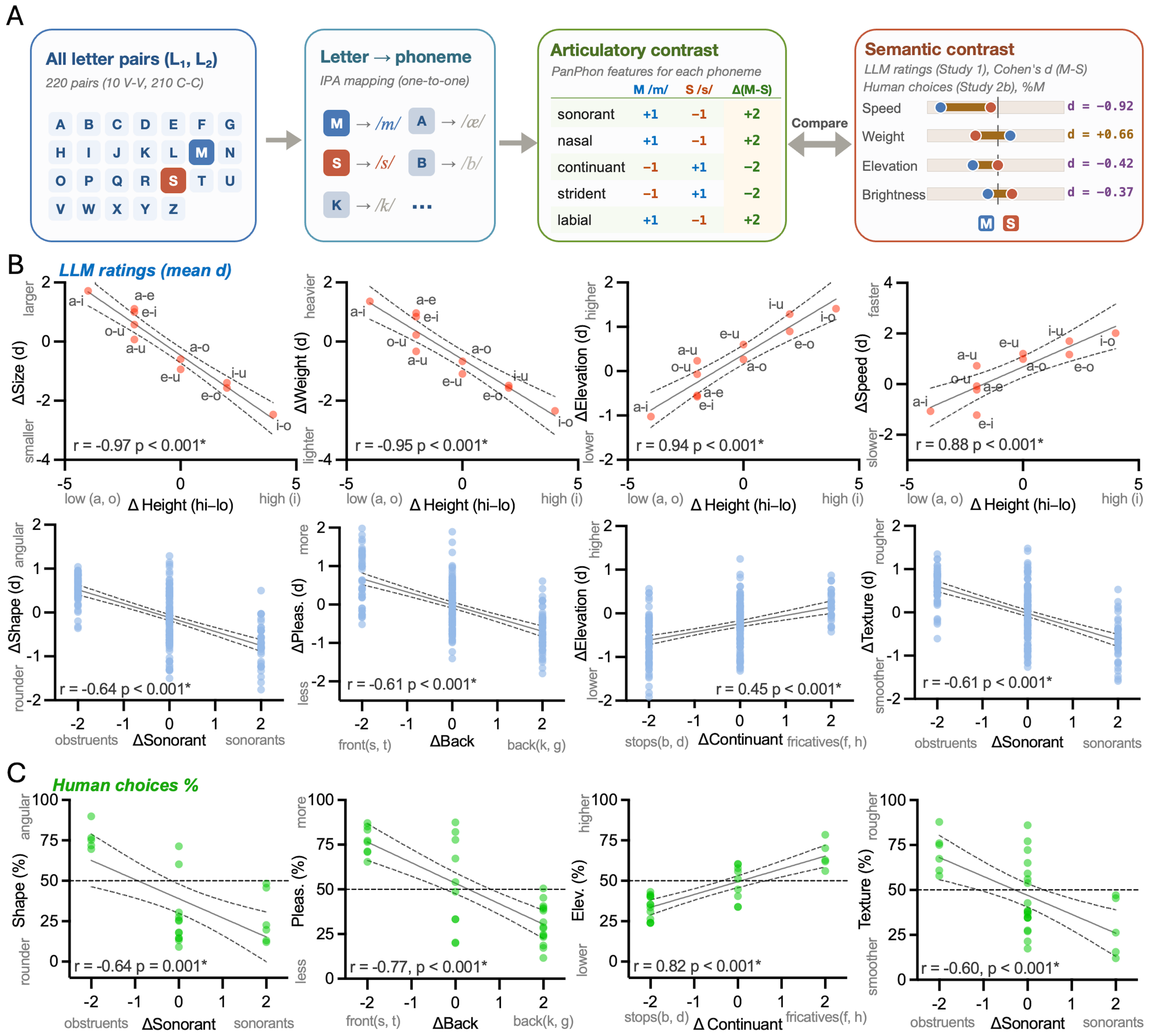

**Figure 4. Articulatory features and letter–meaning associations.**
(A) Analysis pipeline. Each of the 220 letter pairs (10 vowel–vowel, 210 consonant–consonant) is mapped to phonemes (Methods), their articulatory features are taken from PanPhon, and the feature difference between the two letters is compared with the semantic contrast: the model Cohen's d from Study 1, and the percentage of participants choosing the letter-a word in Study 2b.
(B) Model ratings. Top: vowel height (hi − lo) against d for size, weight, elevation, and speed (10 vowel pairs). Bottom: sonorant against shape and texture, back against pleasantness, and continuant against elevation (210 consonant pairs).
(C) The same four consonant relations on the human choices of Study 2b. Each point is one contrast tested on that dimension (15–33 per dimension); y, the percentage choosing the letter-a word, oriented so that the upper pole is spiky, more pleasant, higher, or rougher.
Solid lines, OLS fits with 95% CI; r, Pearson correlation over pairs (letter-level p values in Dataset S4); dashed lines, chance (50%).

## Discussion

Previous work established that particular speech sounds carry particular meanings[3,19–21]. In this study we asked whether this holds across a whole alphabet, and found that each of the 26 English letters carries a structured, multidimensional semantic profile. The profile is consistent across independent language models, confirmed by English readers, perceived by listeners of typologically diverse languages, and predictable from articulatory features.

The patterns we recover align with several classical sound-symbolism findings: the bouba–kiki effect (sonorants as round, obstruents as angular; $r = -0.64$)[9,10,27], Sapir's vowel size ranking (high vowels small, low vowels large; $r = -0.97$)[3,24], and the link between voiceless, sibilant consonants and negative feeling[28] (in our data, sonorants read as more pleasant; $r = 0.59$). These convergences indicate that the multidimensional associations detected here reflect the same form–meaning regularities documented across decades of sound-symbolism research.

Prior work has narrowed the channels that could carry these associations. The shape of the written letter is ruled out by replication across writing systems[14], in early blind individuals[29], and in preliterate infants[30], and hearing is ruled out by deaf and hard-of-hearing participants, who retain sound-symbolic sensitivity and lose it when their articulators are restrained[16]. The experience of producing a sound is the most plausible remaining carrier, and Study 4 is consistent with it: articulatory features predict letter-level semantic profiles in both humans and models. The features predicted a held-out pair well when its two letters appeared in other training pairs (leave-one-pair-out, $r = 0.87$), but less well when a whole letter was held out (leave-one-letter-out, $r = 0.49$). They therefore place a known letter consistently across the pairs it enters, but cannot recover a new letter's position precisely by adding up the effects of its features. Articulatory features predict the associations, but the present study did not manipulate articulation directly[16], so the mechanism itself remains untested.

One caveat is specific to this analysis. The letter-to-phoneme mapping (Methods) is used only to retrieve articulatory features and is an approximation. It assigns /k/ to C, K, Q, and X, so the regression predicts no difference among them. Participants nonetheless told these four letters apart when they were contrasted with one another, so the mapping discards structure that a finer representation might recover.

The nine dimensions serve as representative coordinates of the space of sensation. They are correlated across letters (size with weight, r = 0.95, brightness with elevation, r = 0.85), as they are in the cross-modal literature[17] and in the physical world, so a letter's profile is best read as a position in a correlated space rather than as nine separate scores. Furthermore, two letters can score alike on all nine and still differ on something else, and asking about that something else helps: in Study 3b, the 20 contrasts that had reached only 55.6% in print under a nine-dimension question reached 65.9% in audio under a question written for the pair (Fig. S3B). The comparison is confounded with presentation, so it is suggestive only, but it indicates that the nine dimensions are a frame, not a limit, since what people can feel is not confined to nine axes. Which attribute a given pair of letters differs on is itself an empirical question, of the kind that structure-mapping accounts of iconicity set out to answer[25,26].

Several points bear on how these results should be interpreted. First, the language models in Study 1 are a screening instrument and not a substitute for human judgment, and proprietary models pose a reproducibility problem[31,32], which we address by depositing prompt and raw output and by resting the paper's claims on the preregistered human studies. Second, the associations could in principle be learned from the English lexicon rather than grounded in the sound[12,33]. Our data weigh against this: listeners of five languages who never saw the spelling were above chance, and accuracy did not vary with English proficiency. Third, rare letters carry a flavor of their own, and readers can interpret a nonword through the real words it resembles[34].

Both are properties of written English, and neither accounts for the effects here: they persist with the rarest letters removed and after letter frequency is adjusted for, and they replicate in speech for listeners of four other languages. Fourth, temperature was the weakest dimension in both the model and the human data, plausibly because, unlike size (oral-cavity volume) or shape (constriction type), it has no direct articulatory correlate.

In sum, these findings extend the case for systematic sound–meaning correspondence from selected contrasts to the whole alphabet. One intriguing question is whether letter-level semantic profiles compose at the word level. If each letter carries a structured profile, a word's meaning might partly reflect the superposition of its letters' profiles. Suggestively, "dig" and "drop" are built from letters our data associate with heaviness and downward movement, while "up" and "lift" contain letters associated with lightness and elevation. However, further systematic testing across large lexicons is needed.

## Materials and methods

### Ethics statement

Studies 2a, 2b, 3a, and 3b involved human participants, were conducted in accordance with the Declaration of Helsinki, were approved by the Columbia University Institutional Review Board (IRB-ACYY1798), and were preregistered on the Open Science Framework. All participants gave informed consent, and no personally identifiable information was collected.

### Stimuli and Study 1

Nonword pairs were generated by prompting Claude Opus 4.6 (prompt in Dataset S1). Each pair consists of two nonwords of 4–6 letters that differ in one target letter, occurring once or twice, with all other letters held constant. After screening for real words, unattested clusters, and doubled letters, the pool held 4,420 pairs covering all 220 letter contrasts (210 consonant, 10 vowel) and 8,714 unique nonwords. GPT-4o, Claude Sonnet 4, and Gemini 2.5 Flash Lite rated each nonword five times on nine 0–100 scales (size, shape, brightness, texture, speed, temperature, pleasantness, weight, elevation). For each contrast and dimension, Cohen's d was computed over the pairs sharing the contrast and tested against zero, Benjamini–Hochberg corrected. Neither the sampling temperature nor the rating instruction changes the effect sizes (Dataset S1).

**Study 2.** Native English speakers recruited on Prolific (a nonprobability convenience sample, as in all human studies here) saw a nonword pair and chose the one that better matched a semantic pole. Study 2a ($N = 93$ after preregistered exclusions) tested 54 pairs, six per dimension, selected from Study 1 for large and consistent effects across the three models. Study 2b ($N = 688$) tested all 220 contrasts, each on the dimension with its largest three-model d, with two carrier pairs per contrast drawn at random from the pool, in eight forms of 55 trials.

**Study 3.** Native speakers of English, Spanish, Japanese, Mandarin, and Arabic heard the two nonwords of a pair, synthesized with Microsoft Edge TTS, with the question in their own language. Study 3a ($N = 155$) tested 30 pairs selected from Study 1, each with a question written from a descriptor of the letter and preregistered. Study 3b ($N = 452$) tested 40 contrasts drawn at random from the remaining 190. For 20 of them the pair and dimension question were carried over from Study 2, for the other 20 the pair was drawn at random and the question and predicted direction were written by the author before data collection, without consulting the models.

**Study 4.** Each letter was assigned a canonical phoneme and its 21 PanPhon articulatory features[23]. Consonants and vowels were analyzed separately. For every pair, the feature difference $\Delta f$ was related to the contrast's three-model Cohen's d on each dimension by Pearson correlation, Benjamini–Hochberg corrected, and by linear regression with leave-one-pair-out and leave-one-letter-out cross-validation. Because each consonant enters 20 pairs, tests were repeated at the letter level ($n = 21$) and by permuting letter labels. Letter frequency (SUBTLEX-US, log-transformed) was partialed out of the correlations and added to the regressions.

**Statistical analysis**

Analyses were run in Python 3.10. Multiple comparisons were corrected by the Benjamini–Hochberg procedure ($q < 0.05$). The unit of each test (participant, item, contrast, or letter) is stated with the result. Response times were analyzed with mixed models of log RT on correctness with crossed random intercepts for participants and items.

**Data, Materials, and Software Availability**

Studies 2a, 2b, 3a, and 3b were preregistered on the Open Science Framework. The preregistrations, stimuli, survey code, anonymized participant data, the raw outputs of the three rating models and two open-weight models, and the item-level data (Datasets S1 to S4) will be made available upon publication.

**Acknowledgments**

**Use of AI-Assisted Technologies**

Large language models (GPT-4o, Claude, and Gemini) were used as experimental instruments in Study 1, as described in the Methods. Additionally, AI-assisted tools (Claude, Anthropic) were used to assist with manuscript formatting and copy-editing. The author is fully responsible for the accuracy and integrity of all scientific content, data analysis, interpretation, and writing.

**Funding:** This work was not supported by external funding.

**Author Contributions:** G.Z. conceived and designed the study, developed the experimental paradigm, collected and analyzed all data, and wrote the manuscript.

**Competing Interest Statement:** The author declares no competing interests.

## Supporting Information

**This SI Appendix includes:**

Supporting text (SI Materials and Methods)
Figures S1 to S6
Tables S1 to S5
Legends for Datasets S1 to S4
SI References

**Other supporting materials for this manuscript include the following:**

Datasets S1 to S4 (available upon publication)

## SI Materials and Methods

### Nonword generation

Candidates were screened to exclude real English words, unattested onset and coda clusters, doubled consonants, and identical adjacent vowels. The few items that escaped the screen are flagged in Dataset S1. Because covering all 220 contrasts places some letters where English never does, a minority of items fall outside English orthotactics, and we use "nonword" rather than "pseudoword" throughout. The task instructions and questions shown to participants used "pseudoword" and are reproduced verbatim.

**Study 1:** The models were queried through their APIs. Each nonword was rated five times per model at sampling temperature 0.7, and ratings were averaged across repetitions. Temperature 0.7 was used because the repetitions must vary to estimate rating reliability. At temperature 0 they are identical. Each contrast's d was tested against zero with a one-sample t test.

**Choice of rating models:** The three rating models are proprietary, and their training data and fine-tuning are not disclosed[2]. The models serve as a screening instrument, so the property required of them is that their predictions anticipate human judgment. Against the human choices of Study 2b (440 items), the three-model consensus agrees on 80.8% of items and the individual models on 72.3–78.9%. Two open-weight models run under the same protocol, Llama 3.3 70B and OLMo 2 13B, agree on 65.8% and 56.1%. Llama 3.3 reproduces the letter-level structure (144 of 220 contrasts differentiated on at least one dimension), whereas OLMo 2, whose ratings do not replicate across repetitions (split-half $r = 0.03$), does not. We therefore used the proprietary models for screening and deposit the open-weight ratings.

**Sampling and administration:** All human studies used nonprobability convenience samples recruited through Prolific and were run online in a browser-based interface built by the author. Study 2a was fielded on 28 March 2026, Study 2b on 23–24 August 2026, Study 3a on 15–28 March 2026, and Study 3b on 27 August 2026. On each trial the two response options were the two nonwords of the pair, with left–right position assigned at random, and trial order was randomized for each participant. All questions are given verbatim in Tables S2 to S5 and in full in Datasets S2 and S3. Language models were used to generate and screen the nonwords and, in Study 3a, the candidate question wordings, as described in this section, and every item was reviewed by the author before preregistration. No AI was used in data collection or in scoring responses.

**Study 2:** In Study 2a, 130 native English speakers were recruited via Prolific and 93 were retained after the preregistered exclusions (failed attention checks, reported reading disability or non-native English, or completed the task in under 2 minutes). Pairs were also required to be pronounceable. Three attention checks (real-word size and speed judgments and an instruction-following check) were embedded among the trials. In Study 2b the two carrier pairs of each contrast were drawn independently from the Study 1 pool, excluding pairs used in Study 2a. Forms A–D and E–H each partitioned the 220 contrasts, so the two carrier pairs of a contrast never appeared in the same form, and each form carried three attention checks. Of 810 submissions, 688 were retained under the same exclusions (Dataset S2).

**Study 3:** Participants were recruited via Prolific, 30–35 per language in Study 3a and 100 in Study 3b. Nonwords were synthesized with the en-US-AriaNeural voice, and the same audio was used in all languages. Participants who failed an attention or audio check, finished in under 2 minutes, reported a hearing impairment, or reported a non-target native language were excluded. Of 162 submissions in Study 3a and 500 in Study 3b, 155 and 452 were retained.

The 30 pairs of Study 3a were selected for high cross-model agreement and pronounceability. For each contrast, several wordings written from the same descriptor were put to the three models and the one they agreed on most was kept. The descriptors, the wording presented, and the alternatives tested are in Dataset S3. The 40 contrasts of Study 3b were drawn with no questions or model ratings used in the draw, and all 6 vowel contrasts were included. The 20 carried-over pairs had reached at least 75% human agreement in print. The other 20 questions were written with no model-agreement selection (Dataset S3). Non-native participants rated their spoken English comprehension on a six-level scale. Every Study 3b pair was auditioned before data collection and rejected if the contrast was inaudible or if the audible difference went beyond the contrast letters. The 30 pairs of Study 3a were checked afterward by the same procedure: 23 met it, and excluding the other 7 changes no estimate (Dataset S3).

**Study 4: Articulatory-phonetic prediction**

**Letter–phoneme mapping**. Each of the 26 English letters was assigned a canonical phoneme: A /æ/, B /b/, C /k/, D /d/, E /ɛ/, F /f/, G /g/, H /h/, I /ɪ/, J /dʒ/, K /k/, L /l/, M /m/, N /n/, O /ɑ/, P /p/, Q /k/, R /ɹ/, S /s/, T /t/, U /ʌ/, V /v/, W /w/, X /k/, Y /j/, Z /z/. The mapping is used only to convert letters into articulatory features and is an approximation: vowels take their short values, as in the closed syllables that make up most of the stimuli, and C, K, Q, and X share /k/. Alternative readings (c → /s/, q → /kw/, x → /ks/ or /s/) change the correlation between the articulatory prediction and the human choices of Study 2b by at most 0.06 (Dataset S4). Articulatory features were taken from PanPhon[1], which encodes 24 phonological features as +1/−1/0; three that do not apply to English (velaric, hitone, hireg) were excluded, leaving 21.

**Pair-level design.** All 210 consonant–consonant (CC) and 10 vowel–vowel (VV) pairs were enumerated. The semantic contrast of a pair on each dimension is its Cohen's d from Study 1, computed separately for each model and then averaged across the three.

**Feature filtering.** Two filters were applied in sequence to the starting 21 features. Features that take a single value across all letters in a subset carry no information. Within the 21 consonants, six features (syl, sg, cg, lo, tense, long) are constant and were removed, leaving 15 CC features (son, cons, cont, delrel, lat, nas, strid, voi, ant, cor, distr, lab, hi, back, round). Within the 5 vowels, sixteen features are constant, leaving 5 VV features (delrel, hi, lo, back, tense). These yielded the $15 \times 9 = 135$ CC and $5 \times 9 = 45$ VV univariate Pearson correlations (with the vowel-height composite hi − lo additionally displayed, yielding 54 total VV cells) displayed in Fig. S5A.

**Rank filter (Fig. S5B regression and Fig. S5C $\Delta R^2$ decomposition).** Among the 15 CC features, four (strid, distr, back, round) are exact linear combinations of the remaining eleven within the 21-consonant subset (residuals $< 10^{-14}$; e.g., strid = 1 − son + cont + delrel + nas). Among the 5 VV features, tense is an exact linear combination of the remaining four (tense = −1 − 2·delrel + lo + back). Including any of these features would render the design matrix rank-deficient. The regression models therefore retained 11 CC features (son, cons, cont, delrel, lat, nas, voi, ant, cor, lab, hi) and 4 VV features (delrel, hi, lo, back).

**Feature classes (Fig. S5C).** For the unique-contribution analysis, features were grouped following the SPE feature-geometry tradition into manner (son, cons, cont, delrel, lat, nas, strid), place (ant, cor, distr, lab, hi, lo, back, round), and laryngeal (voi, sg, cg). Following standard phonological practice, the tongue-body features hi, lo, and back were grouped under "place" because they encode dorsal place of articulation in consonants (e.g., velars are +hi +back); for vowels, the same features encode height and backness, but because the consonant and vowel models were fit separately this dual role does not introduce ambiguity. After applying the rank filter, the classes effectively entering the CC regression were manner = {son, cons, cont, delrel, lat, nas}, place = {ant, cor, lab, hi}, laryngeal = {voi}; for the VV regression, place = {hi, lo, back} with delrel as the only manner feature. $\Delta R^2$ for each class was computed as $R^2$(full model) − $R^2$(model with that class removed). For the vowel panels, height was scored as hi − lo: +2 for high vowels (I, U), 0 for mid (E), −2 for low (A, O); Δ across the 10 vowel pairs takes values in {−4, −2, 0, +2, +4}. All other panels use a single PanPhon feature. For each panel, r is the Pearson correlation between Δ feature and Δ Cohen's d across the 210 consonant pairs or 10 vowel pairs.
**Statistical inference.** At the letter level, a letter's score is its mean effect across its 20 pairs (n = 21). Leave-one-letter-out cross-validation removes the 20 pairs containing that letter. It cannot be run on the vowels, since removing one of five leaves six pairs, fewer than the regression has parameters.

**Statistical analysis**

Analyses used SciPy, NumPy, pandas, and scikit-learn, and figures were made in GraphPad Prism 10 and matplotlib. Cohen's d for a letter contrast is the mean rating difference between paired nonwords divided by the standard deviation of those differences across the pairs sharing the contrast. Split-half reliability was computed by splitting the carrier pairs of each contrast at random into two halves, computing d in each, and correlating the profiles, over 100 splits, with Spearman–Brown correction. The permutation null for Study 1 reassigned word pairs to contrasts at random, keeping the number of pairs per contrast fixed, and re-ran the full pipeline 1,000 times; a sign-flip variant, which keeps the pairing and randomizes the direction of each pair's difference, gave the same result (Dataset S1).

Letter frequency was taken from SUBTLEX-US and log-transformed. Where it is used as a covariate, the difference in log frequency between the two letters of a contrast was signed toward the predicted letter, and the adjusted value reported is the intercept of the regression of the human effect on that difference, that is, the effect expected of two equally frequent letters.

In Study 2a, overall accuracy and each dimension were tested with one-sample t tests on the 93 participant accuracies, and each item with a two-sided exact binomial test against .50. In Study 2b, the confirmatory test was a percentile bootstrap confidence interval on the mean of the 220 contrast accuracies, resampling participants within form (10,000 replicates); items were tested by exact binomial test and dimensions by one-sample t test on item accuracies; agreement across bands of model effect size was compared by one-way ANOVA. In Studies 3a and 3b, each contrast was tested with a two-sided exact binomial test pooled across languages; for each language group we report the median participant accuracy with a distribution-free 95% confidence interval and treat the group as above chance where the interval excludes 50; English proficiency levels were compared by one-way ANOVA. Correlations between model effect and human agreement, between carrier replicates, between print and audio, and between language groups are two-sided Pearson correlations over the units named in the figure legends; the ten between-language correlations were Benjamini–Hochberg corrected.

Response-time models were fit separately for each study, and effects are reported as $\beta$ with the equivalent percentage $100 \times (1 - e^{\beta})$. Responses under 200 ms, over 30 s, or negative were excluded (26 of 5,022 trials in Study 2a, 158 of 37,840 in 2b, 52 of 4,650 in 3a, and 590 of 18,080 in 3b); every estimate is unchanged without this exclusion. The short responses are all in Study 3b, whose clock starts when the second clip ends and so measures decision time only, whereas in Study 3a it starts when the trial appears and includes playback; absolute response times are therefore not comparable between the two. The participant-level comparisons in Figs. S2F and S4A use all trials: consistent versus inconsistent trials by paired t test on each participant's median response time, and trial position by paired t test on the fourth minus the first quartile of trials, on log response time.

## Supplementary Figures

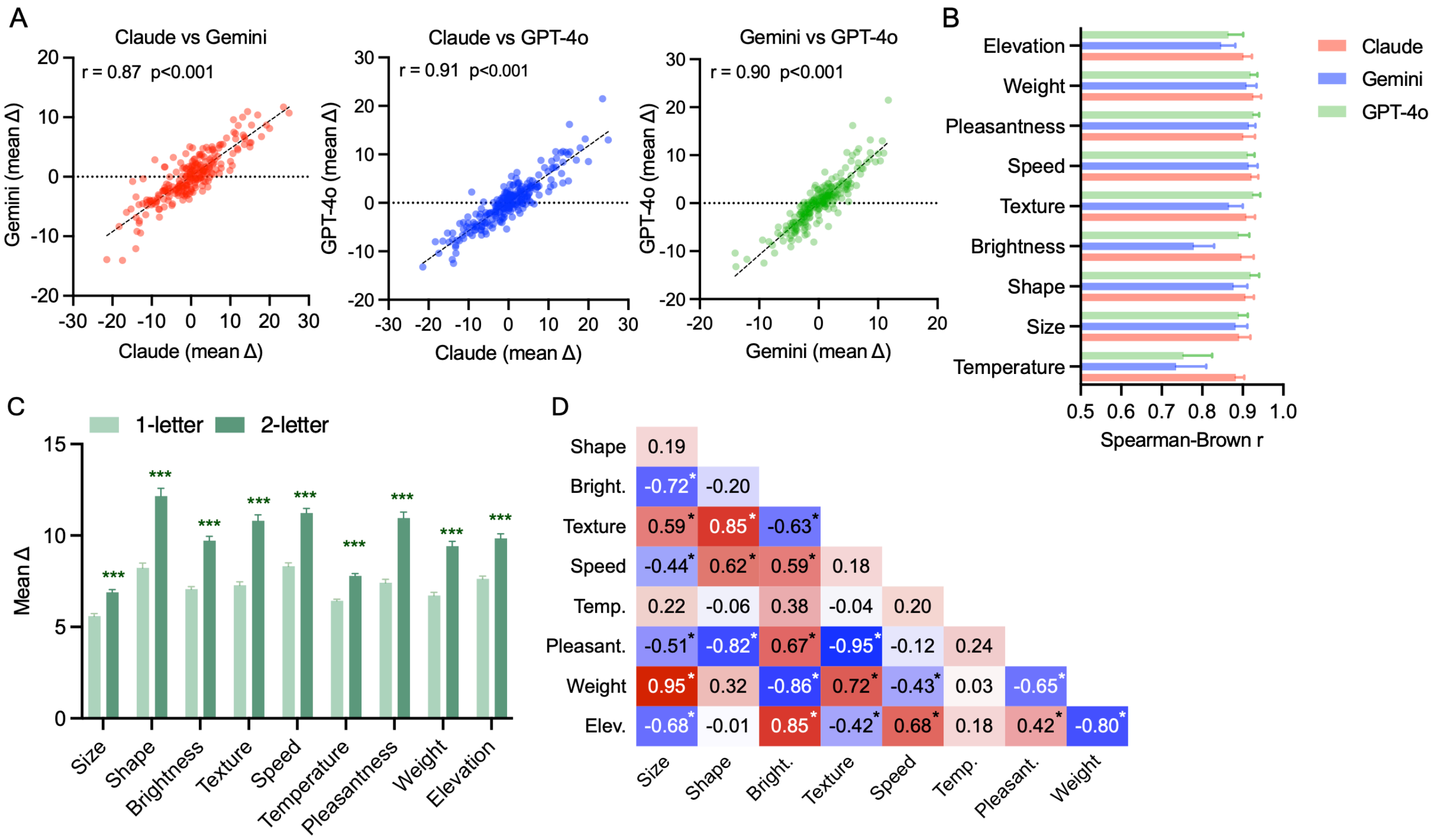

**Figure S1. Cross-model agreement, internal consistency, dosage effect, and inter-dimension structure.**

(A) Cross-model agreement. Each point is one letter on one dimension (26 letters × 9 dimensions = 234 points); axes, the letter's mean rating difference (Δ, the word containing the letter minus its partner, averaged over all pairs containing the letter) in each model.

(B) Split-half reliability. For each model and dimension, the nonword pairs of every contrast were randomly split into two halves, the mean Δ of each half was computed, and the two halves were correlated across the 220 contrasts. Bars, mean over 100 random splits, Spearman–Brown corrected; error bars, minimum and maximum over the 100 splits.

(C) Dosage effect. Mean |Δ| (0–100 scale) for pairs in which the target letter occurs once (light, 1,700 pairs) or twice (dark, 2,720 pairs). For each of the 220 contrasts, |Δ| was averaged over the three models and over the pairs at each dose; bars, mean over contrasts; error bars, SEM over contrasts.

(D) Inter-dimension correlations. Pearson r among the nine dimensions across the 26 letter-level Cohen's d profiles (three-model mean), the full matrix from which Fig. 1C is drawn.

In (A), Pearson r and p are shown in each panel. In (C), *** $p < 0.001$, paired t test over the 220 contrasts. In (D), * $p < 0.05$, two-sided test of Pearson r (n = 26 letters), uncorrected.

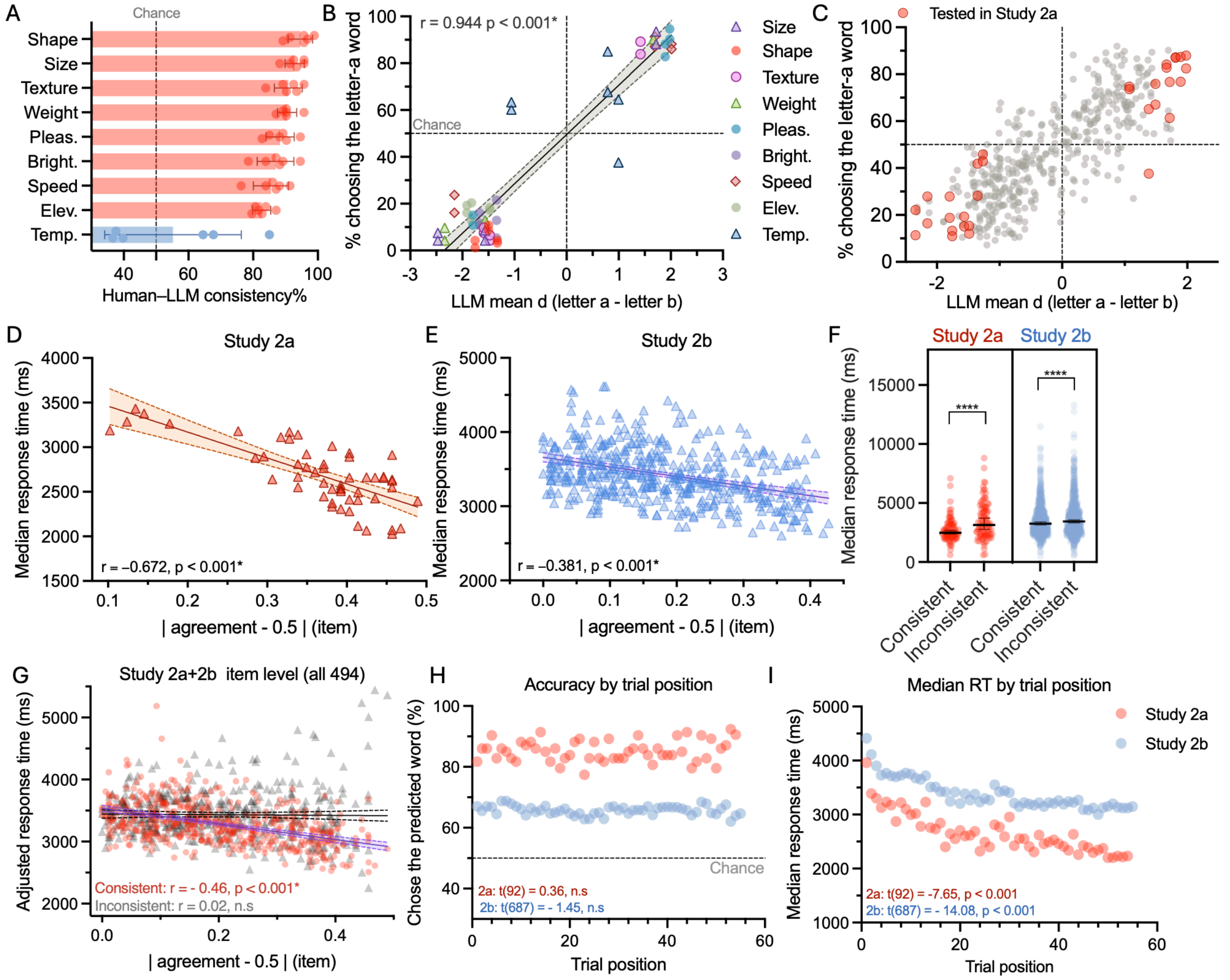

**Figure S2. Studies 2a and 2b: item-level agreement, response times and trial position.**
(A) Study 2a accuracy by dimension. Points, the 54 items (six per dimension); bars, mean with 95% CI.
(B) Study 2a, 54 items: the model effect (three-model mean d, letter a minus letter b) against the percentage of participants choosing the letter-a word, colored by dimension.
(C) Study 2b, all 440 items, plotted as in (B); red, contrasts also tested in Study 2a.
(D, E) Median response time per item against item agreement (|agreement − 0.5|) in Study 2a (D) and Study 2b (E).
(F) Per-participant median response time on trials consistent and inconsistent with the predicted direction, Study 2a ($n = 93$) and Study 2b ($n = 688$); lines, median with 95% CI.
(G) Item-level participant-adjusted response time against item agreement for responses consistent (red points, purple line) and inconsistent (gray points, black line) with the predicted direction, Studies 2a and 2b pooled (494 items).
(H, I) Accuracy (H) and median response time (I) by trial position in Study 2a (red) and Study 2b (blue); each point is one trial position, averaged over participants.
Solid lines, OLS fits with 95% CI; r, Pearson. (B) $r = 0.94$, (D) $r = -0.67$, (E) $r = -0.38$, (G) consistent $r = -0.46$, all $p < 0.001$; (G) inconsistent $r = 0.02$, n.s. (F) **** $p < 0.0001$, paired t test. (H, I) paired t tests on the fourth minus the first quartile of trials within participants, t and p shown in each panel.

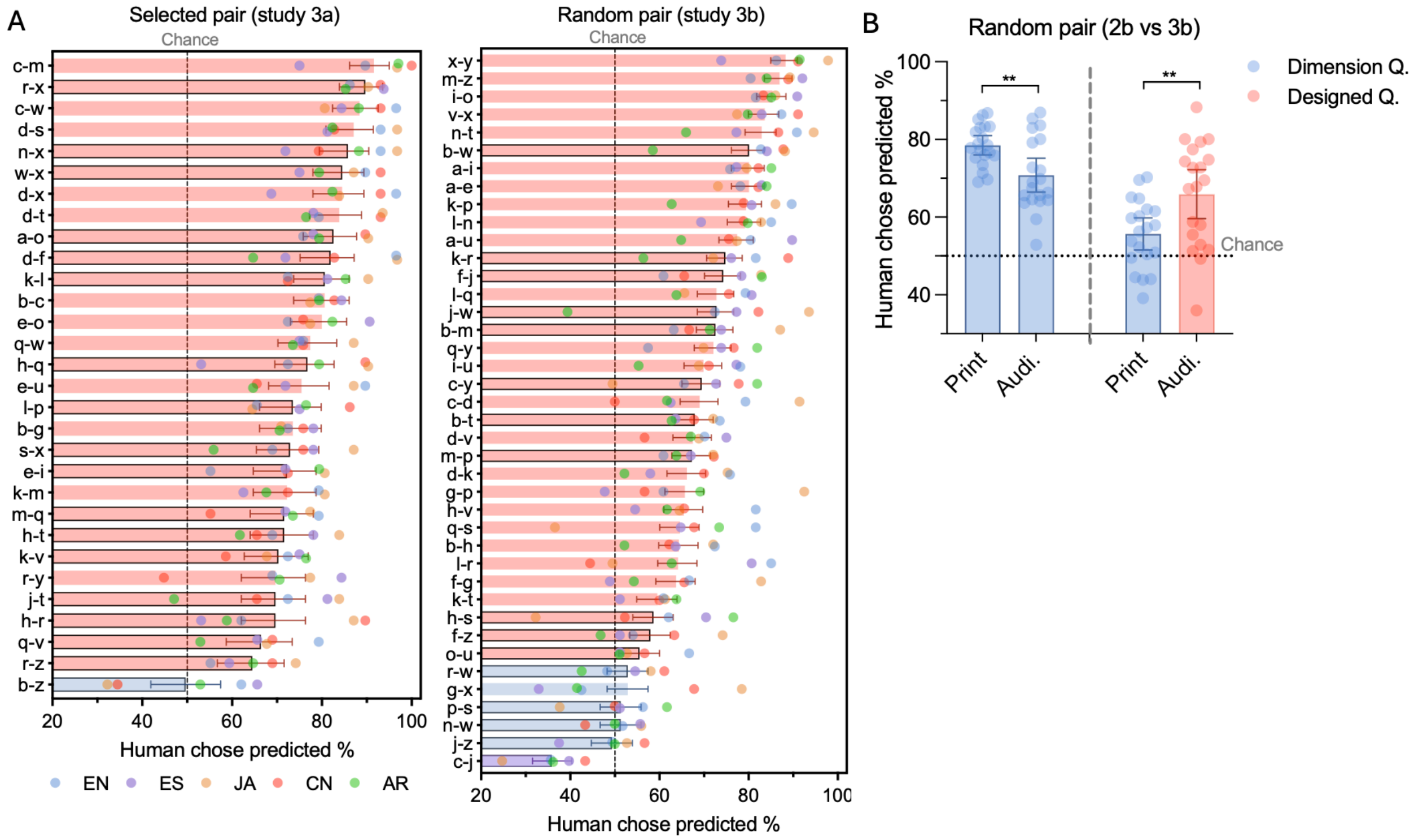

**Figure S3. Per-contrast accuracy in the auditory studies and the effect of question type.**
(A) Percentage choosing the predicted word for each contrast of Study 3a (30 selected pairs, left) and Study 3b (40 randomly drawn contrasts, right). Points, the five language groups; bars, mean of the five groups with 95% CI; outlined bars, contrasts asked a designed question (all of Study 3a and 20 of Study 3b); blue, contrasts not significantly above chance; purple, significantly reversed (c–j); dashed line, chance.
(B) Study 3b contrasts by question type. Blue, the 20 contrasts that kept the pair and dimension question from Study 2a or 2b; red, the 20 asked a designed question written for the audio study. Print, accuracy for the same contrast in Study 2b (dimension question); audio, Study 3b. Points, contrasts; bars, mean with 95% CI.
In (A), each contrast was tested against chance with a two-sided binomial test over all its trials (Study 3a, n = 155; Study 3b, n = 452), Benjamini–Hochberg corrected within each study; red, significant above chance; blue, not significant; purple, significantly below chance.

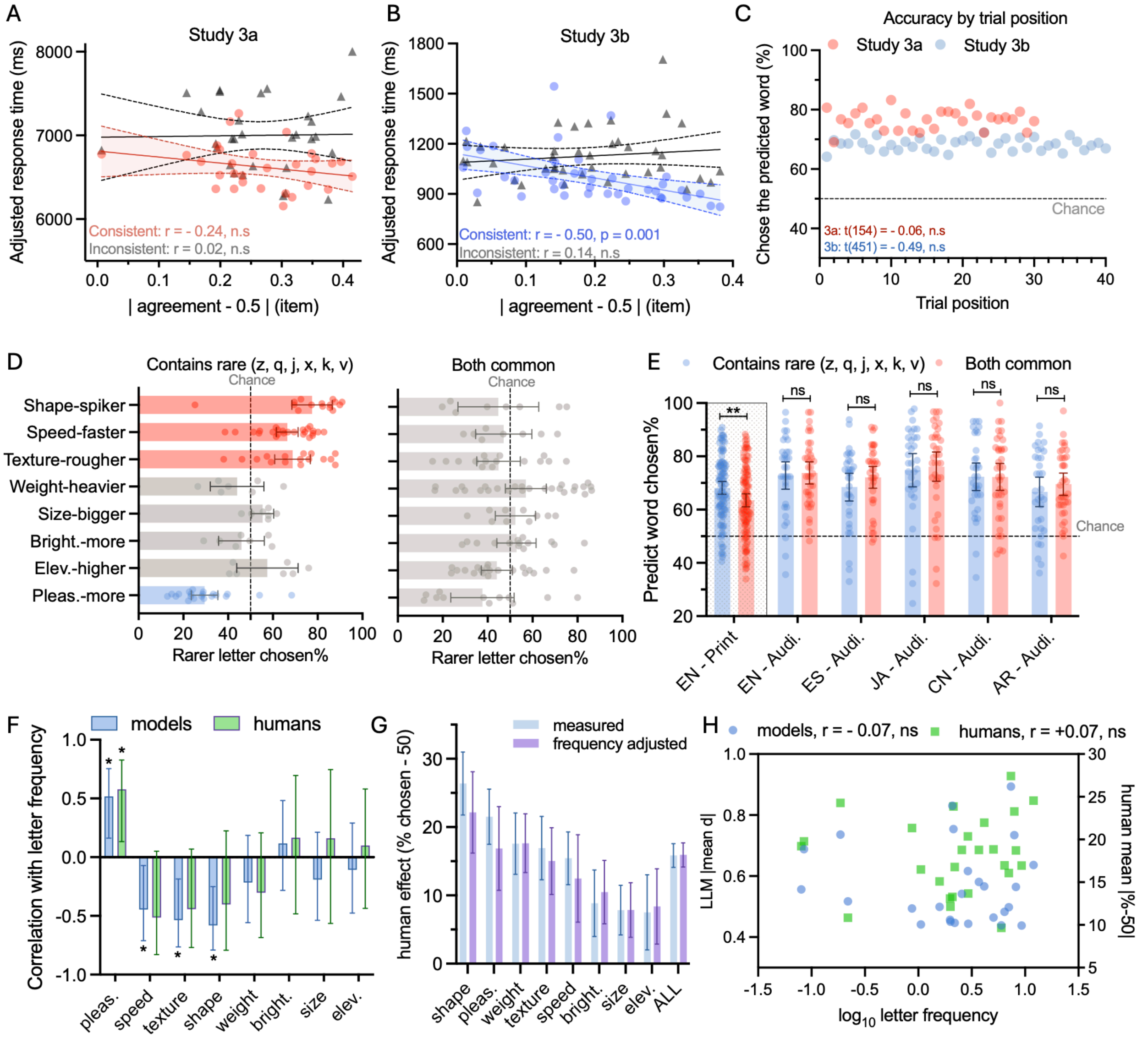

**Figure S4. Response times and trial position in the auditory studies, and letter frequency.** **(A-B)** Item-level participant-adjusted response time against item agreement (|agreement − 0.5|) for responses consistent (colored) and inconsistent (gray) with the predicted direction, Study 3a (A, 30 items) and Study 3b (B, 40 items). In Study 3a the clock starts when the trial appears and includes audio playback, in Study 3b when the second clip ends, so the two scales are not comparable. **(C)** Accuracy by trial position in Studies 3a and 3b, averaged over participants. **(D)** Study 2b: percentage of trials on which the word carrying the rarer of the two letters was chosen for the pole named on the axis, for contrasts containing one of the six rarest letters (z, q, j, x, k, v; left) and for contrasts of two common letters (right). Red/blue, above/below chance; gray, n.s. Temperature (two contrasts) omitted. **(E)** Accuracy for contrasts containing a rare letter (blue) and for contrasts of two common letters (red), in print (Study 2b) and in audio (Studies 3a and 3b). **(F)** Pearson correlation across letters between letter frequency and the letter-level profile. **(G)** Human effect per dimension in Study 2b (percentage points from chance) as measured and after adjustment for the difference in log frequency between the two letters of each contrast. **(H)** Mean absolute effect of each letter against its frequency, for the models (left axis) and humans (right axis).

Points in (D, E), contrasts; bars, mean with 95% CI (F, r with 95% CI); solid lines, OLS fits with 95% CI; r, Pearson (A, B, F, H, n = 26 letters in H). (C) paired t tests on the fourth minus the first quartile of trials within participants. (D) one-sample t tests against 50% over contrasts, Benjamini–Hochberg q < 0.05. (E) unpaired t tests. (F) * p < 0.05.

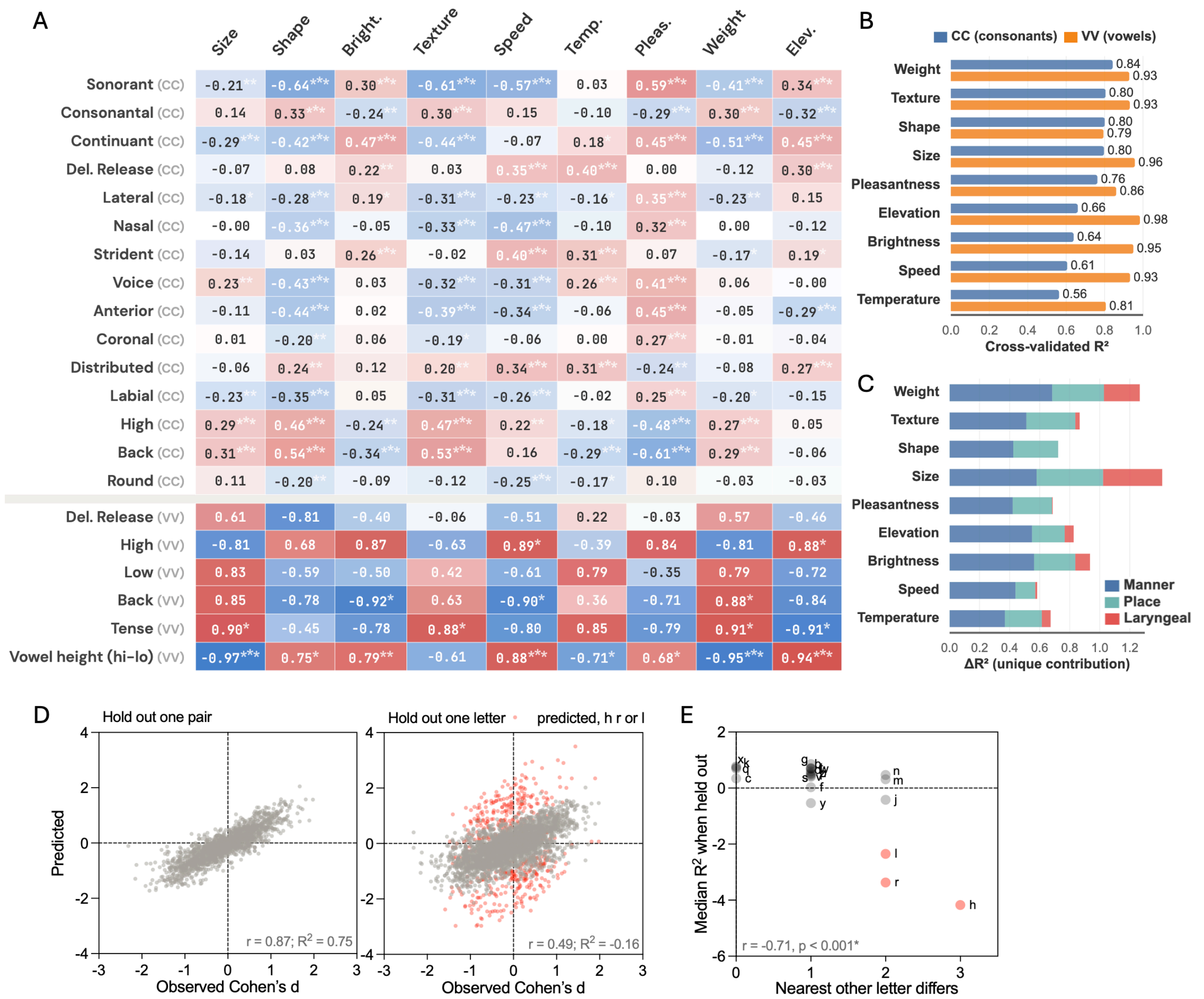

**Figure S5. Articulatory feature analyses on the model ratings.**
(A) Pearson r between the feature difference (Δf) and the three-model Cohen's d across letter pairs, for consonant pairs (15 features × 9 dimensions, 210 pairs) and vowel pairs (5 features plus the vowel-height composite hi − lo, 10 pairs). Red, positive; blue, negative. Letter-level tests are in Dataset S4.
(B) Leave-one-pair-out cross-validated $R^2$ by dimension for the consonant (CC, 210 pairs, 11 features) and vowel (VV, 10 pairs, 4 features) models. With five vowels and four features almost any feature set fits, so the vowel values are not independent evidence.
(C) Unique contribution of each feature class to the consonant model, $\Delta R^2 = R^2$(full model) − $R^2$(model with that class removed), stacked by class.
(D) Predicted against observed d for every consonant pair on every dimension when one pair (left) or one whole letter (right) is held out. Red, folds in which the held-out letter is h, r, or l.
(E) Median cross-validated $R^2$ across the nine dimensions when each consonant letter is held out, against the number of features separating it from its nearest other consonant; red, h, r, and l.
(A) asterisks, Benjamini–Hochberg corrected over the cells of each panel, * $q < 0.05$, ** $q < 0.01$, *** $q < 0.001$. (D) r, Pearson correlation between predicted and observed d over the 1,890 pair × dimension values; $R^2$, cross-validated $R^2$ over the same values, negative when the predictions fit worse than the mean. (E) r, Pearson over the 21 consonants.

A

| | Human participants | | | | | | | | LLM (on the human' scale) | | | | | | | |
| --- | --- | --- | --- | --- | --- | --- | --- | --- | --- | --- | --- | --- | --- | --- | --- | --- |
| | Size | Shape | Bright. | Texture | Speed | Pleas. | Weight | Elev. | Size | Shape | Bright. | Texture | Speed | Pleas. | Weight | Elev. |
| pairs asked | 15 | 23 | 20 | 30 | 31 | 33 | 29 | 27 | 15 | 23 | 20 | 30 | 31 | 33 | 29 | 27 |
| **Sonorant** (CC) | .21 | -.64** | .42 | -.60** | -.36 | .75*** | -.11 | .45 | .08 | -.75*** | .07 | -.55* | -.55* | .81*** | -.06 | .44 |
| **Consonantal** (CC) | -.13 | .32 | -.16 | .06 | .13 | -.24 | .02 | -.10 | .19 | .33 | .03 | .21 | .02 | -.48* | .04 | -.34 |
| **Continuant** (CC) | .29 | -.34 | .42 | -.64** | .58** | .74*** | -.59** | .82*** | .12 | -.17 | -.02 | -.53* | .16 | .70*** | -.37 | .72*** |
| **Del. Release** (CC) | -.19 | .11 | -.21 | .13 | .37 | — | -.03 | -.14 | -.05 | .06 | .23 | .06 | .61** | — | -.10 | .14 |
| **Lateral** (CC) | — | .09 | .61* | -.43 | -.05 | .59** | -.19 | .32 | — | -.09 | .44 | -.17 | -.21 | .45* | .02 | .23 |
| **Nasal** (CC) | -.25 | -.61** | -.19 | -.31 | -.34 | .43 | .14 | -.39 | .27 | -.69** | .03 | -.33 | -.48* | .41 | -.00 | -.38 |
| **Strident** (CC) | -.11 | -.10 | -.16 | -.20 | .80*** | .18 | -.45 | .29 | .10 | .09 | .10 | -.18 | .67*** | .03 | -.39 | .32 |
| **Voice** (CC) | .48 | -.52* | -.28 | -.40 | .16 | .57** | .62** | -.22 | .49 | -.64** | -.03 | -.33 | -.03 | .61** | .57* | -.13 |
| **Anterior** (CC) | -.04 | -.75*** | -.25 | -.42 | .11 | .64*** | -.35 | -.28 | .24 | -.64** | -.31 | -.37 | -.34 | .50* | -.21 | -.49* |
| **Coronal** (CC) | -.01 | -.21 | -.39 | -.28 | .54** | .64*** | -.15 | -.25 | .46 | -.20 | -.12 | -.22 | .24 | .48* | .21 | -.15 |
| **Distributed** (CC) | -.13 | .33 | .13 | .35 | -.06 | -.64*** | .14 | .12 | -.50 | .27 | .33 | .23 | .40 | -.48* | -.26 | .22 |
| **Labial** (CC) | -.14 | -.64** | -.04 | -.24 | -.34 | .12 | -.09 | -.12 | -.56 | -.56* | -.26 | -.30 | -.17 | .13 | -.49* | -.26 |
| **High** (CC) | .15 | .74*** | .29 | .58** | -.44 | -.70*** | .42 | -.09 | -.20 | .65** | .02 | .43 | -.12 | -.54** | .44 | .23 |
| **Back** (CC) | .16 | .68** | .16 | .37 | -.42 | -.77*** | .50* | .32 | -.07 | .59* | -.20 | .36 | -.13 | -.75*** | .41 | .35 |
| **Round** (CC) | .23 | -.32 | -.12 | -.06 | -.08 | .17 | .24 | -.06 | -.03 | -.33 | -.45 | -.21 | -.02 | .21 | .26 | .12 |

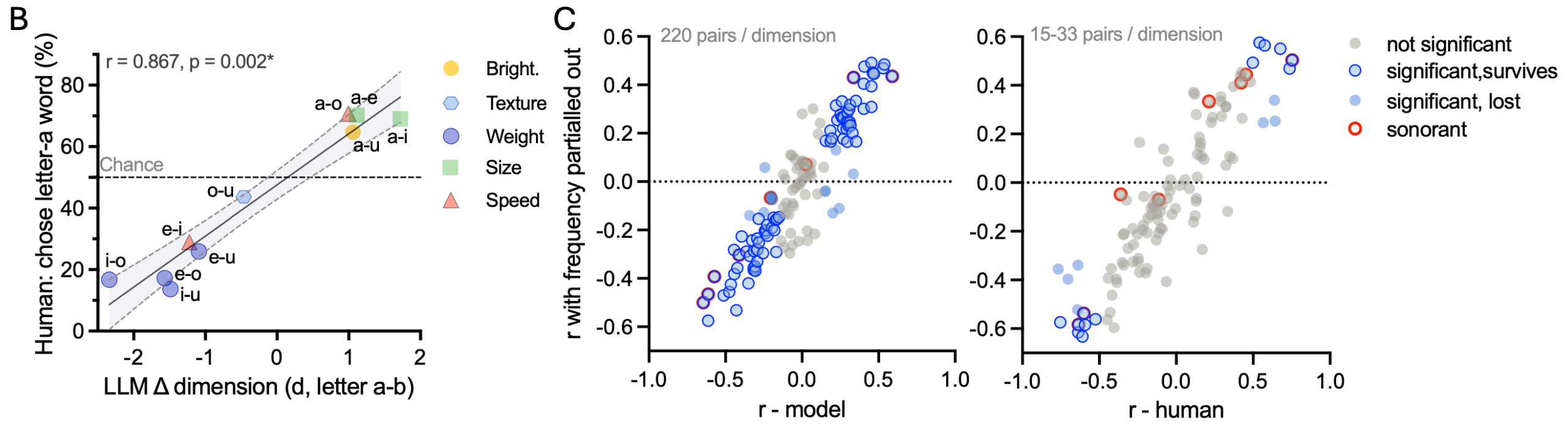

**Figure S6. The articulatory analysis on human choices.**
(A) Feature × dimension correlations for consonant contrasts, computed on the human choices of Study 2b (left) and on the models' choices between the same two carrier words, expressed as a percentage (right). The top row gives the number of contrasts tested on each dimension; temperature had only two contrasts and was excluded from this analysis. Dashes, the feature does not vary among the contrasts tested. Across the 118 filled cells the two matrices correlate at r = 0.84, and all 33 cells significant in either matrix agree in sign.
(B) Vowel contrasts: the model effect (d, letter a minus letter b) against the percentage of participants choosing the letter-a word in Study 2b, colored by dimension. Solid line, OLS fit with 95% CI.
(C) Each point is one consonant feature × dimension cell: x, the correlation as in Fig. S5A (models, left, 135 cells, 210 pairs per dimension) or in (A) (humans, right, 118 cells, 15–33 contrasts per dimension); y, the same correlation after the difference in log letter frequency between the two letters of each pair is partialed out of both variables. Blue outline, significant before and after adjustment; blue fill, significant before but not after; gray, not significant before adjustment; red outline, the sonorant row.
All r, Pearson. (A) * q < 0.05, ** q < 0.01, *** q < 0.001, Benjamini–Hochberg corrected within each matrix. (C) significance at Benjamini–Hochberg q < 0.05 within each panel.

**Table S1. Semantic dimensions and rating scales used in Study 1.**
Nine bipolar dimensions were rated on a 0–100 scale. Each nonword was rated five times per model (GPT-4o, Claude Sonnet 4, Gemini 2.5 Flash Lite), and ratings were averaged across repetitions.

| Dimension | Score |
|---|---|
| size | size: 0 = very small, 100 = very large |
| shape | shape: 0 = very round, 100 = very spiky |
| brightness | brightness: 0 = very dark, 100 = very bright |
| texture | texture: 0 = very smooth, 100 = very rough |
| speed | speed: 0 = very slow, 100= very fast |
| temperature | temperature: 0 = very cold, 100 = very hot |
| pleasantness | pleasantness: 0 = very unpleasant, 100 = very pleasant |
| weight | weight: 0 = very light, 100 = very heavy |
| elevation | elevation: 0 = very low position, 100 = very high position |

**Table S2. Representative forced-choice stimuli and results for Study 2a.**
Predicted = nonword predicted by LLM consensus to match the semantic pole; Accuracy = proportion of participants selecting the predicted nonword (chance = 50%). Full stimulus set available in Dataset S2. N = 93.

| Dimension | Question | Pair | Predicted | Alternative | Accuracy |
|---|---|---|---|---|---|
| Shape | Which feels spikier? | x_d | efvoxa | efvoda | 89.2% |
| | | x_m | laxbix | lambim | 98.9% |
| Size | Which feels bigger? | o_i | brob | brib | 92.5% |
| | | o_e | ogmo | egme | 95.7% |
| Weight | Which feels heavier? | o_i | fomro | fimri | 90.3% |
| | | o_i | blopt | blipt | 95.7% |
| Texture | Which feels rougher? | x_h | oeviax | oeviah | 90.3% |
| | | q_l | qiepqa | liepla | 95.7% |
| Pleasantness | Which feels more pleasant? | h_x | ahpohe | axpoxe | 88.2% |
| | | l_q | taleul | taqeuq | 94.6% |
| Brightness | Which feels brighter? | y_d | bityoy | bitdod | 87.1% |
| | | y_d | puyeay | pudead | 94.6% |
| Speed | Which feels faster? | i_u | brif | bruf | 91.4% |
| | | z_m | ziszoe | mismoe | 83.9% |
| Elevation | Which feels higher? | y_m | uyray | umram | 82.8% |
| | | y_d | puyeay | pudead | 87.1% |
| Temperature | Which feels warmer? | y_l | meuyot | meulot | 39.8% |
| | | a_i | rasrad | risrid | 67.7% |

**Table S3. Summarized results for Study 2b.**

Study 2b extends to the complete set of 220 letter contrasts, each contrast with two random-drawn word pairs (440 items). Full item-level results are in Dataset S2. N = 688.

| Dimension | Question | Contrasts | Items | Accuracy | 95% CI |
|---|---|---|---|---|---|
| Shape | Which feels spikier? | 23 | 46 | 76.4% | [73.1, 79.7] |
| Size | Which feels bigger? | 17 | 34 | 57.8% | [54.6, 61.1] |
| Texture | Which feels rougher? | 31 | 62 | 66.9% | [63.5, 70.4] |
| Weight | Which feels heavier? | 33 | 66 | 67.6% | [64.3, 70.8] |
| Pleasantness | Which feels more pleasant? | 33 | 66 | 71.5% | [68.3, 74.8] |
| Brightness | Which feels brighter? | 21 | 42 | 58.8% | [55.0, 62.6] |
| Speed | Which feels faster? | 33 | 66 | 65.4% | [62.4, 68.4] |
| Elevation | Which feels higher? | 27 | 54 | 57.5% | [53.5, 61.5] |
| Temperature | Which feels warmer? | 2 | 4 | 53.7% | [43.1, 64.4] |

**Table S4. Study 3a auditory stimuli and results.** (Full results available in Dataset S3)
Row represents one letter pair presented as synthesized audio to speakers of five languages. Predicted Word = nonword predicted by LLM consensus.

| Pair | Predicted Word | Other Word | Question (Dimension) | Average% (N=155) |
|---|---|---|---|---|
| c_m | acuci | amumi | Which feels more angular/sharp? | 91.7% |
| r_x | xuxseu | rurseu | Which feels more like sudden cracking? | 89.7% |
| c_w | acucie | awuwie | Which feels more angular/sharp? | 88.6% |
| d_s | dirft | sirft | Which feels heavier? | 87.2% |
| n_x | xomexa | nomena | Which feels more like violent cracking? | 85.9% |
| d_x | efxaxe | efdade | Which feels more angular/sharp? | 84.9% |
| w_x | fuoxex | fuowew | Which feels more like structural destruction? | 84.9% |
| d_t | umtat | umdad | Which feels faster? | 84.1% |
| a_o | molvoz | malvaz | Which feels more like sinking? | 82.7% |
| d_f | fufeol | dudeol | Which feels more floating? | 82.5% |
| b_c | cicfe | bibfe | Which feels more angular/sharp? | 80.7% |
| k_l | bulil | bukik | Which feels more continuously unbroken? | 80.3% |
| o_e | brov | brev | Which feels heavier? | 79.7% |
| q_w | doqaqi | dowawi | Which feels more angular/sharp? | 77.5% |
| h_q | qauqie | hauhie | Which feels more rigid/stiff? | 77.0% |
| e_u | zerem | zurum | Which feels brighter? | 75.8% |
| b_g | gigoa | biboa | Which feels more angular/sharp? | 73.6% |
| l_p | tipipa | tilila | Which feels more forceful? | 73.5% |
| s_x | xexufa | sesufa | Which feels more like destroying integrity? | 73.2% |
| k_m | akuak | amuam | Which feels more angular/sharp? | 72.5% |
| i_e | bleg | blig | Which feels more area-like? | 71.9% |
| h_t | tubitu | hubihu | Which feels more focused/pointing? | 71.6% |
| m_q | quqroz | mumroz | Which feels more piercing? | 71.5% |
| h_r | erear | eheah | Which feels more like turning/revolving? | 70.2% |
| k_v | itoev | itoek | Which feels more fluidly overflowing? | 70.0% |
| j_t | putut | pujuj | Which feels more straight and pointing? | 70.0% |
| r_y | iperar | ipeyay | Which feels heavier? | 69.2% |
| q_v | vavoun | qaqoun | Which feels more like flowing outward? | 66.9% |
| r_z | zeilzi | reilri | Which feels more like confrontation? | 64.5% |
| b_z | zieszu | biesbu | Which feels more interlacing? | 49.5% |

**Table S5. Study 3b auditory stimuli and results.** (Full results available in Dataset S3)
Predicted Word = nonword predicted by Study 2 human consensus (dimension questions), or by the author (designed questions). N=452.

| Pair | Predicted Word | Other Word | Question | Average% |
|---|---|---|---|---|
| x–y | exiuxo | eyiuyo | Which feels more angular/sharp? | 88.3% |
| m–z | ziszoe | mismoe | Which feels faster? | 87.0% |
| i–o | blopt | blipt | Which feels bigger? | 85.4% |
| v–x | rotnax | rotnav | Which feels spikier? | 83.6% |
| n–t | etlet | enlen | Which feels spikier? | 83.0% |
| a–e | apad | eped | Which feels bigger? | 80.1% |
| a–i | smas | smis | Which feels bigger? | 80.1% |
| b–w | bunbop | wunwop | Which feels more like bouncing back? | 80.1% |
| k–p | kaeku | paepu | Which feels spikier? | 79.4% |
| l–n | lilit | ninit | Which feels brighter? | 79.2% |
| a–u | apab | upub | Which feels brighter? | 77.4% |
| k–r | rorlo | koklo | Which feels more rolling and continuous? | 74.8% |
| f–j | aufeo | aujeo | Which feels more like a gentle stroke? | 74.3% |
| j–w | woawe | joaje | Which feels fuller and rounder? | 72.8% |
| l–q | qiepqa | liepla | Which feels rougher? | 72.8% |
| b–m | mopma | bopba | Which feels wrapped in and embraced? | 72.6% |
| q–y | yoypia | qoqpia | Which feels more pleasant? | 72.1% |
| i–u | brif | bruf | Which feels faster? | 69.9% |
| c–y | tiyou | ticou | Which feels lighter and finer? | 69.5% |
| c–d | cuceo | dudeo | Which feels brighter? | 69.0% |
| b–t | eitoa | eiboa | Which feels more directional? | 67.9% |
| d–v | suvit | sudit | Which feels higher? | 67.5% |
| m–p | puepoa | muemoa | Which feels more rigid/sturdy? | 67.3% |
| d–k | kikbi | didbi | Which feels spikier? | 66.2% |
| g–p | augeg | aupep | Which feels heavier? | 65.7% |
| h–v | atvov | athoh | Which feels spikier? | 65.5% |
| q–s | zenqoq | zensos | Which feels rougher? | 64.6% |
| b–h | uehio | uebio | Which feels higher? | 64.4% |
| l–r | nurad | nulad | Which feels rougher? | 64.2% |
| f–g | govge | fovfe | Which feels heavier? | 63.7% |
| k–t | ikako | itato | Which feels spikier? | 59.5% |
| h–s | huehu | suesu | Which feels more dispersing? | 58.6% |
| f–z | urziov | urfiov | Which feels more constricting? | 58.0% |
| o–u | dopod | dupud | Which feels more like settling down? | 55.5% |
| g–x | exipab | egipab | Which feels faster? | 52.9% |
| r–w | wuwio | rurio | Which feels more monotonous? | 52.9% |
| n–w | piduna | piduwa | Which feels more like stalling? | 51.3% |
| p–s | opapa | osasa | Which feels more rigid/sturdy? | 51.3% |
| j–z | josjed | zoszed | Which feels more like darting away? | 49.3% |
| c–j | focoes | fojoes | Which feels more like wrapping around? | 35.8% |

## Legends for Datasets S1 to S4

**Dataset S1 (available upon publication).** Study 1 stimuli and model ratings. The 4,420 nonword pairs and 8,714 nonwords covering all 220 letter contrasts, the prompt used to generate them, the rating prompt, the raw ratings of the three models on the nine dimensions, Cohen's d for every contrast and dimension, the temperature and instruction sensitivity check, and the permutation null.

**Dataset S2 (available upon publication).** Studies 2a and 2b. Item-level accuracy, per-participant summaries and trial-level data for both studies, the letter-frequency analyses, and the response-time mixed models.

**Dataset S3 (available upon publication).** Studies 3a and 3b. Per-pair accuracy overall and by language group, per-participant summaries, trial-level data, exclusions, the letter descriptors and question wordings of Study 3a, the questions and predicted directions of Study 3b, the audition check, self-rated English comprehension, and the response-time models.

**Dataset S4 (available upon publication).** Study 4. The letter-to-phoneme mapping and PanPhon features of each letter, feature differences and effect sizes for the 210 consonant and 10 vowel pairs, feature × dimension correlations on the model ratings and on the human choices, letter-frequency partialling, letter-level tests, cross-validation, and the mapping sensitivity analysis.

## SI References